\documentclass[10pt,twocolumn,letterpaper]{article}

\usepackage[pagenumbers]{cvpr} % To force page numbers, e.g. for an arXiv version
\usepackage{amsthm}
\usepackage{algorithm}
\usepackage{algorithmic}
\newcommand{\methodName}{CaliTex}

\definecolor{cvprblue}{rgb}{0.21,0.49,0.74}
\usepackage[pagebackref,breaklinks,colorlinks,allcolors=cvprblue]{hyperref}

\def\confName{CVPR}
\def\confYear{2026}

\title{CaliTex: Geometry-Calibrated Attention for View-Coherent \\ 3D Texture Generation}

\author{%
\hspace{-.7em}%
Chenyu Liu$^{1}$,\;
Hongze Chen$^{2}$,\;
Jingzhi Bao$^{3}$,\;
Lingting Zhu$^{4}$,\;
Runze Zhang$^{5}$,\;
Weikai Chen$^{5}$,\;
Zeyu Hu$^{5}$,\;\\[0.1em]
Yingda Yin$^{5}$,\;
Keyang Luo$^{5}$,\;
Xin Wang$^{5}$ \\[0.2em]
$^{1}$PKU, $^{2}$HKUST, $^{3}$CUHK(SZ), $^{4}$HKU, $^{5}$LIGHTSPEED
}

\begin{document}
%\maketitle

\twocolumn[{%
\renewcommand\twocolumn[1][]{#1}%
\maketitle
\vspace{-24pt}
\begin{center}
    \centering
    \captionsetup{type=figure}
    \includegraphics[width=\linewidth]{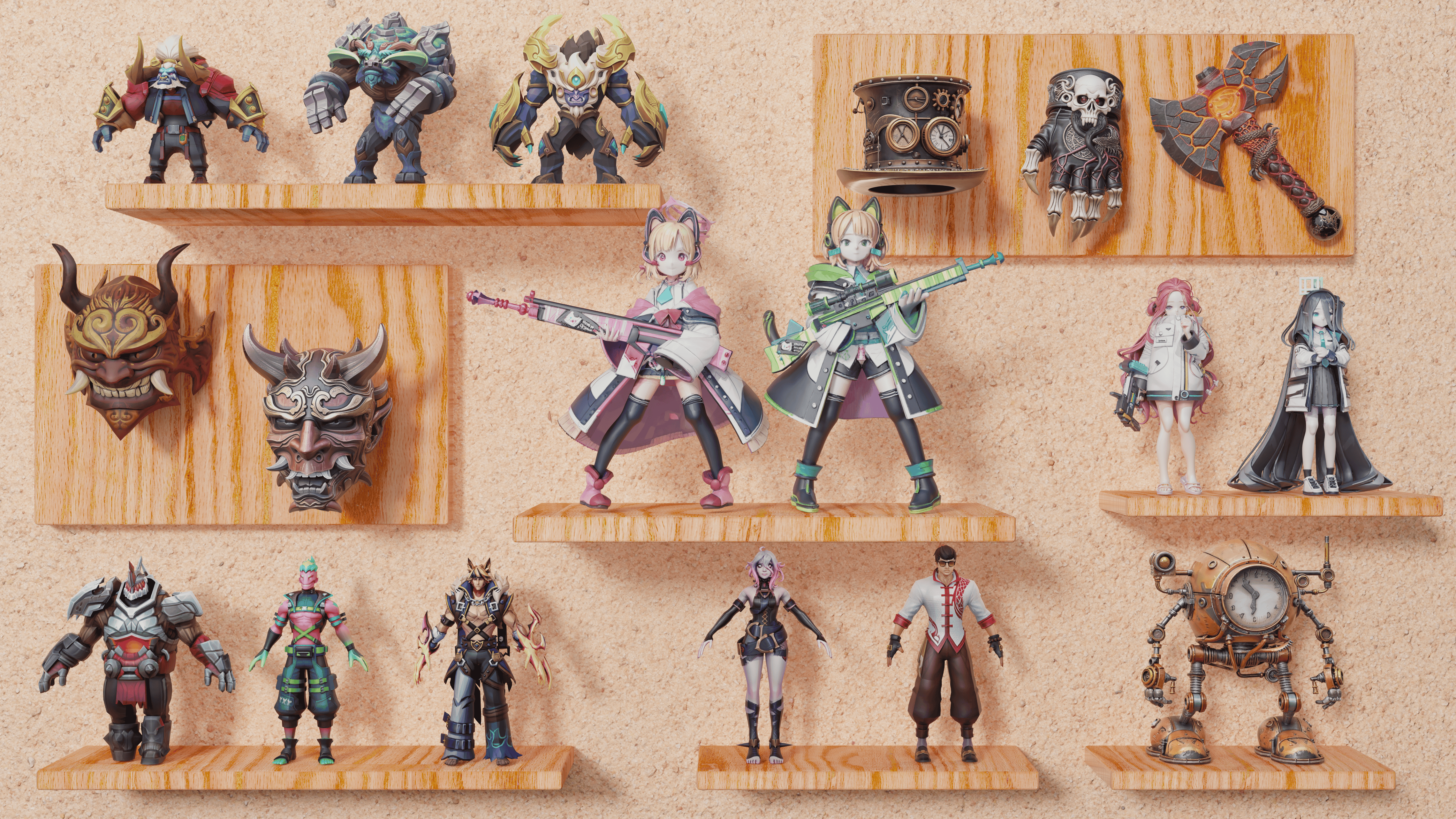}
    \vspace{-1.2em}
    \captionof{figure}{A collection of 3D objects textured by our method, demonstrating high-fidelity, seamless and geometry-aligned textures facilitated by our framework with geometry-calibrated attention. Visit our project website at \tt \href{https://calitex-project.github.io}{https://calitex-project.github.io}.}
    \label{fig:teaser}
\end{center}%
}]

\newif\ifdrafting
\draftingtrue
% \draftingfalse
\ifdrafting
    \newcommand{\jz}[1]{\textcolor{cyan}{[JZ: #1]}}
\else
    \newcommand{\jz}[1]{}
\fi

\ifdrafting
    \newcommand{\cy}[1]{\textcolor{red}{[CY: #1]}}
\else
    \newcommand{\cy}[1]{}
\fi

\ifdrafting
    \newcommand{\rz}[1]{\textcolor{orange}{[RZ: #1]}}
\else
    \newcommand{\rz}[1]{}
\fi

\begin{abstract}
Despite major advances brought by diffusion-based models, current 3D texture generation systems remain hindered by cross-view inconsistency -- textures that appear convincing from one viewpoint often fail to align across others. We find that this issue arises from attention ambiguity, where unstructured full attention is applied indiscriminately across tokens and modalities, causing geometric confusion and unstable appearance-structure coupling.
To address this, we introduce \methodName{}, a framework of geometry-calibrated attention that explicitly aligns attention with 3D structure.
It introduces two modules: Part-Aligned Attention that enforces spatial alignment across semantically matched parts, and Condition-Routed Attention which routes appearance information through geometry-conditioned pathways to maintain spatial fidelity.
Coupled with a two-stage diffusion transformer, \methodName{} makes geometric coherence an inherent behavior of the network rather than a byproduct of optimization.
Empirically, \methodName{} produces seamless and view-consistent textures and outperforms both open-source and commercial baselines. 
\end{abstract}
% Recent developments in multi-view diffusion models have demonstrated their ability to generate high-quality 3D model textures.
% However, models pre-trained with full attention often present ambiguous attention patterns: 
% (1) Geometry details cheat, and cross-view tokens are ambiguously associated according to geometrically similar but distinct regions;
% (2) Conditions confuse, and noise tokens alternate unsteadily in attention between geometry and reference features.
% These ambiguities lead to mis-aligned local details and geometry-texture inconsistencies, causing artifacts after projection.
% To address this issue, we introduce two attention mechanisms that explicitly disentangle and guide attention flows:
% (1) 3D Local Attention, which constrains token interactions within each semantic part to reduce across-view ambiguity, and (2) Condition-Bridged Attention, which regulates attention between noise and reference tokens via geometry conditions to enhance alignment.
% Extensive experiments demonstrate that our method effectively resolves attention ambiguity, producing high-fidelity 3D textures with consistent geometry alignment and fine-grained details, outperforming all existing open-source and commercial baselines.    
\begin{figure*}
    \centering
    \includegraphics[width=0.97\linewidth]{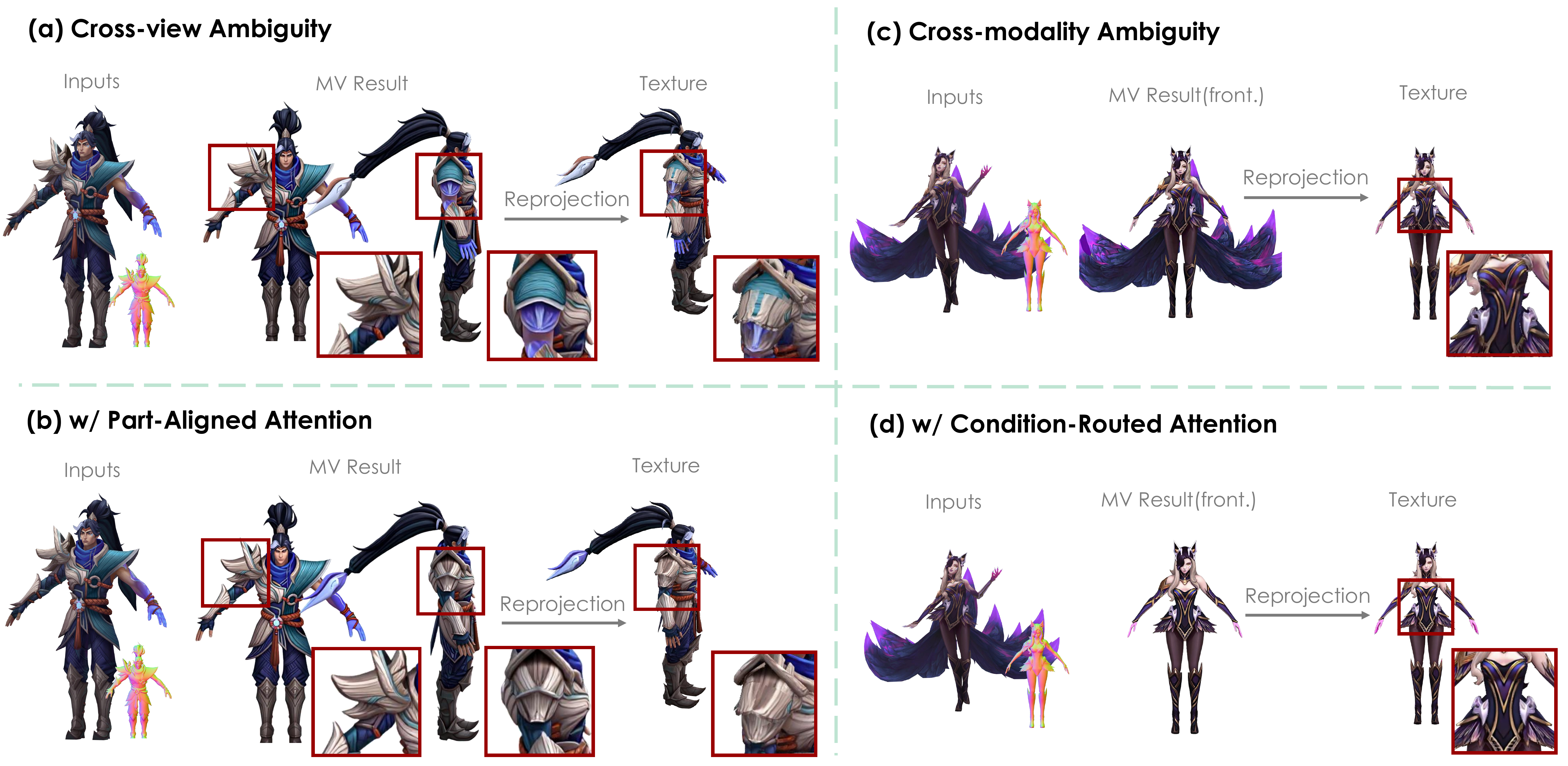}
    \caption{\textbf{Illustration issues caused by attention ambiguity and our proposed solutions.}
Zoom in for more details. 
(a) The model confuses the left limb in the second view with the right limb, producing seams in the texture.
(b) Our Part-Aligned Attention constrains attention computation within semantic parts, effectively eliminating cross-view inconsistency.
(c) The model directly copies visually similar regions from the reference image, leading to misalignment with the geometry condition.
(d) Our Condition-Routed Attention ensures geometry-aligned texture generation, correcting the distortion on the clothing, as highlighted in the bottom-right.
}
    \vspace{-0.6em}
    \label{fig:intro_show}
\end{figure*}

\section{Introduction}
\label{sec:intro}

The automatic generation of 3D assets holds immense promise for content creation in gaming, film, and virtual reality.
Recent advances in native 3D generative modeling~\cite{lai2025hunyuan3d25highfidelity3d, yuan2025seqtexgeneratemeshtextures, li2025step1x,zhang2024clay} have significantly improved generation quality, marking a shift from traditional SDS-based pipelines~\cite{poole2022dreamfusion,wang2023prolificdreamer} to more powerful diffusion-based approaches.
% leading to the gradual replacement of traditional SDS-based methods with more powerful diffusion-based approaches.
Most advanced techniques adopt a two-stage paradigm: geometry is first synthesized, followed by texture generation conditioned on the geometry.
In this process, 2D generative priors from image diffusion models~\cite{rombach2022high,podell2023sdxl,esser2024sd3, blackforest2024flux, zhang2023adding} are leveraged to synthesize multi-view images~\cite{shi2023mvdream,long2024wonder3d,zhang2024clay}, which are then reprojected back onto the surface to construct textures.
While this strategy delivers impressive appearance quality, it often collapses under cross-view inconsistency -- the same surface region can appear differently across generated views, producing seams or blurring after reprojection.
Shown in Fig.~\ref{fig:intro_show} (a,c), such artifacts do not stem from rendering, but rather from a deeper representation misalignment within the model itself.

Unlike conventional text-to-image models, where full attention across all noise and text tokens is both sufficient and necessary to capture a semantic context, texture generation involves a far more intricate interplay among noise tokens, geometry condition tokens, and reference image tokens.
% involve a more complex attention computation among noise tokens, geometry condition tokens, and reference image tokens. 
% (Since an image contains richer appearance information than a text prompt, we focus primarily on the single-image-to-3D generation setting in this work.)
% \rz{explain this complexity?}
Current state-of-the-art methods~\cite{liang2025UnitTEX, zhao2025hunyuan3d, zhu2025muma} simply extend full attention indiscriminately across all tokens and views, presuming correspondences will emerge. However, this premise does not hold in practice.

Such a na\"{i}ve full-attention design leads to two inherent ambiguities that compromise texture consistency and detail.
(i) \emph{Cross-view Ambiguity} (Fig.~\ref{fig:intro_show} (a)). 
Tokens belonging to geometrically similar yet distinct regions, such as the left and right limbs, often attend to each other across views. Consequently, the model confuses distinct parts and generates nearly identical local textures for both. 
After weighted blending, these misaligned views are projected back onto inconsistent surface regions, resulting in seams and loss of spatial coherence.
(ii) \emph{Cross-modality Ambiguity} (Fig.~\ref{fig:intro_show} (c)). 
The model also suffers from unstable cross-modal attention: noise tokens alternatively attend to the reference image or the geometry condition.
This results in either appearance overfitting (copying visual patterns) or geometric overreliance (losing appearance fidelity), producing superficially realistic but geometrically inconsistent textures.

We contend that geometric consistency does not arise automatically from training -- it requires architectural calibration.
Instead of introducing additional supervision or handcrafted priors, we reformulate attention itself to be geometry-aware -- guiding the model to know \emph{where to focus} and \emph{how information flows} across modalities.
Following this principle, we introduce \methodName{}, a framework of \emph{Geometry-Calibrate Attention} that explicitly aligns attention with 3D structure at both spatial and informational levels.
This calibration is instantiated through two complementary mechanisms (Fig.~\ref{fig:intro_show} (b,d)): \emph{Part-Aligned Attention} and \emph{Condition-Routed Attention}.
    
Part-Aligned Attention (PAA) mitigates cross-view ambiguity by imposing geometry-aware locality. The key idea is to constrain attention propagation within the same semantic part, preventing tokens from mistakenly attending to geometrically similar but distinct regions across views.
Specifically, we use PartField~\cite{liu2025partfieldlearning3dfeature} to decompose the mesh into semantic components, assigning each face a discrete part index.
By rendering part-colored images from multiple viewpoints, we obtain part labels for every token, enabling us to group tokens that represent the same 3D part across different views.
Cross-view attention is then computed within each group, while intra-view full attention is preserved to capture broader context.
As shown in Fig.~\ref{fig:intro_show} (b), our method effectively eliminates cross-view inconsistency, yielding aligned fine details across different viewpoints.

Condition-Routed Attention (CRA) addresses cross-modality ambiguity by restructuring how information flows between reference and noise tokens.
Instead of allowing noise and reference tokens to interact freely, CRA imposes a dual-pathway design: one branch computes attention between condition and reference tokens, capturing geometric priors from appearance cues, while the other links condition and noise tokens, injecting those geometry-aware features back into the generative process.
This routing mechanism (Fig.~\ref{fig:intro_show} (d)) ensures that appearance information is always mediated by geometry, suppressing direct reference copying and maintaining consistent texture alignment with the underlying 3D surface.

% To resolve this ambiguity, we propose a \textbf{Condition-Bridged Attention Mechanism} that explicitly regulates the attention flow through the geometry condition.
% Instead of allowing direct interaction between noise and reference tokens, our dual-pathway design enforces a structured attention process: one pathway computes attention between condition tokens and reference tokens to inject geometric priors, while the other pathway links noise tokens with condition tokens to propagate this geometric-aligned information into the generative process.
% As shown in Fig.~\ref{fig:intro_show} (d), this bridged attention structure eliminates ambiguous dependencies on the reference image and ensures that texture generation remains consistently guided by geometric conditions.

Together, these mechanisms are implemented within a two-stage diffusion transformer: a single-view DiT first captures intra-view semantics under full attention, while a multi-view DiT equipped with PAA and CRA enforces cross-view and cross-modal coherence.
This design not only stabilizes attention dynamics but also transforms geometric consistency from a post-hoc regularization into an intrinsic property of the generative process.
Empirically, it yields seamless and geometrically faithful textures, even for complex assets exhibiting high symmetry or self-occlusion, as demonstrated in Fig.~\ref{fig:intro_show} (b,d).  
Our main contributions are summarized as follows:

\begin{itemize}
\item We present \methodName{}, a novel framework of geometry-calibrated attention that embeds geometric reasoning directly into attention computation.
\item We propose part-aligned attention to enforce spatial alignment across views via part-level geometric priors.
\item We introduce conditioned-routed attention to regulate reference-noise interaction through geometry-conditioned routing, achieving stable cross-modal alignment.
\item We achieve high-fidelity 3D texture generation results that surpass existing state-of-the-art baselines.
\end{itemize}

% \item We introduce 3D Local Attention, leveraging geometric priors to enhance cross-view texture consistency.
% \item We propose a Condition-Bridged Attention Mechanism that enhances the geometric alignment of multi-view diffusion models.
% \item We achieve high-fidelity 3D texture generation results that surpass existing open-source and commercial baselines.

% Through these two modules, our method effectively addresses the attention ambiguity issues present in multi-view generation models.
% Extensive experiments demonstrate that our method produces highly detailed textures on complex assets, while significantly reducing visual artifacts. Our approach achieves state-of-the-art performance across all open-source and commercial baselines.
\section{Related Work}
\label{sec:RelatedWork}

\subsection{Texture Generation for 3D Objects}

With the advent of diffusion-based generative models, texture generation for 3D objects has been extensively explored. Early approaches based on Score Distillation Sampling (SDS)~\cite{poole2022dreamfusion, youwang2024paintit, lin2023magic3d, zhang2024dreammat, deng2024flashtex, chen2023fantasia3d, chen2024textto3dusinggaussiansplatting} attempted to optimize 3D representations such as NeRF and 3D Gaussian Splatting ~\cite{mildenhall2020nerfrepresentingscenesneural, kerbl20233dgaussiansplattingrealtime} by rendering images with sampled random noise and predicting the noise using a pre-trained 2D diffusion model. However, SDS methods often lead to oversaturated colors and suffer from the Janus problem. Recently, the rapid progress of text-conditioned image generation models~\cite{esser2024sd3, labs2025flux1kontextflowmatching, lipman2023flowmatchinggenerativemodeling} has inspired many works on 3D texture generation. Some methods~\cite{chen2023text2tex, richardson2023texturetextguidedtexturing3d, zeng2024paint3d} iteratively generate single-view images from arbitrary viewpoints, but often suffer from poor cross-view consistency. More recent approaches generate multiple views simultaneously, demonstrating strong quality~\cite{cheng2024mvpaint, huang2025mv, li2025step1x, hunyuan3d2025hunyuan3d2.1, liang2025UnitTEX, zhu2025muma, yuan2025seqtexgeneratemeshtextures, bao2025lumitex}. However, they fail to achieve proper cross-view and view-geometry alignment when the geometries are highly complex.

Another line of work directly predicts 3D representations with colors~\cite{xiang2025structured3dlatentsscalable, yu2023texturegeneration3dmeshes, yu2024texgen, 10656692, xiong2025texgaussiangeneratinghighqualitypbr}.  UniTEX~\cite{liang2025UnitTEX} introduces a two-stage pipeline: it first generates multi-view images with a diffusion model, followed by a transformer-based Large Texturing Model to predict the texture function in 3D space from both images and geometry. These approaches can inherently avoid issues caused by self-occlusion, but due to the scarcity of large-scale textured 3D training data, their outputs generally underperform compared to methods that leverage existing 2D image generation models in quality.

\subsection{3D Texture Alignment}

The cross-view and view-geometry misalignments significantly affect texture quality, and several recent works have attempted to address these issues. SeqTex~\cite{yuan2025seqtexgeneratemeshtextures} directly generates UV-space textures alongside multi-view images, replacing the traditional back-projection pipeline. However, recent high-quality 3D generation models~\cite{li2025sparc3dsparserepresentationconstruction, chen2025ultra3defficienthighfidelity3d, lai2025hunyuan3d25highfidelity3d} typically rely on automatic UV unwrapping tools such as xatlas~\cite{xatlas}, where fragmented UV layouts often lead to degraded generation quality. Elevate3D~\cite{Ryu_2025} refines single-view generations iteratively to improve cross-view consistency and modifies geometry to maintain coherence with synthesized views, though this may inadvertently alter original geometric details. Romantex~\cite{feng2025romantex} introduces a 3D-aware RoPE to inject spatial information into the generative model and employs a geometry-related CFG technique to enhance alignment with geometry. AlignTex~\cite{zhang2025aligntex} strengthens geometry alignment by fusing image and geometry features within the diffusion process.

%Moreover, due to the limited model capacity, multi-view diffusion models often generate images with relatively low resolution. Hunyuan3D-Paint and MV-Adapter~\cite{zhao2025hunyuan3d, huang2025mv} employ Real-ESRGAN~\cite{wang2021realesrgantrainingrealworldblind} to enhance the resolution of multi-view images before baking them into textures. However, applying super-resolution independently to each view reduces cross-view consistency. Elevate3D~\cite{Ryu_2025} addresses this with an iterative view-by-view refinement strategy using their improved HFS-SDEdit method based on SDEdit~\cite{meng2022sdeditguidedimagesynthesis}. It achieves strong generative quality while maintaining cross-view consistency by incorporating a coarse 3D texture as an additional condition. However, it alters the geometry to maintain consistency with the generated view, which may inadvertently change some of the original geometric details.

\subsection{3D Semantic Segmentation}

Recent progress in image segmentation, especially with models such as SAM~\cite{Kirillov_2023_ICCV}, has also driven advances in 3D semantic segmentation. Several methods use 2D segmentation priors from SAM to improve 3D semantic understanding~\cite{liu2025partfieldlearning3dfeature, tang2025segmentmesh, kim2024garfieldgroupradiancefields, ying2023omniseg3domniversal3dsegmentation, ye2024gaussiangroupingsegmentedit}. Among them, PartField~\cite{liu2025partfieldlearning3dfeature} trains a network with contrastive learning to predict a continuous 3D feature field, which is then clustered to obtain high-quality 3D segmentation efficiently. In contrast, P$^3$-SAM~\cite{ma2025p3samnative3dsegmentation} operates purely in the 3D domain, addressing the spatial inconsistency problem found in methods that depend on 2D segmentation labels. Several works have also explored part-level generation~\cite{yang2025omnipartpartaware3dgeneration, ding2025fullpartgenerating3dresolution} and part-level refinement in 3D generation tasks~\cite{chen2025ultra3defficienthighfidelity3d}, which inspires us to incorporate semantic part priors into the 3D texture generation.
\section{Method}

\begin{figure*}
    \centering
    \includegraphics[width=0.95\linewidth]{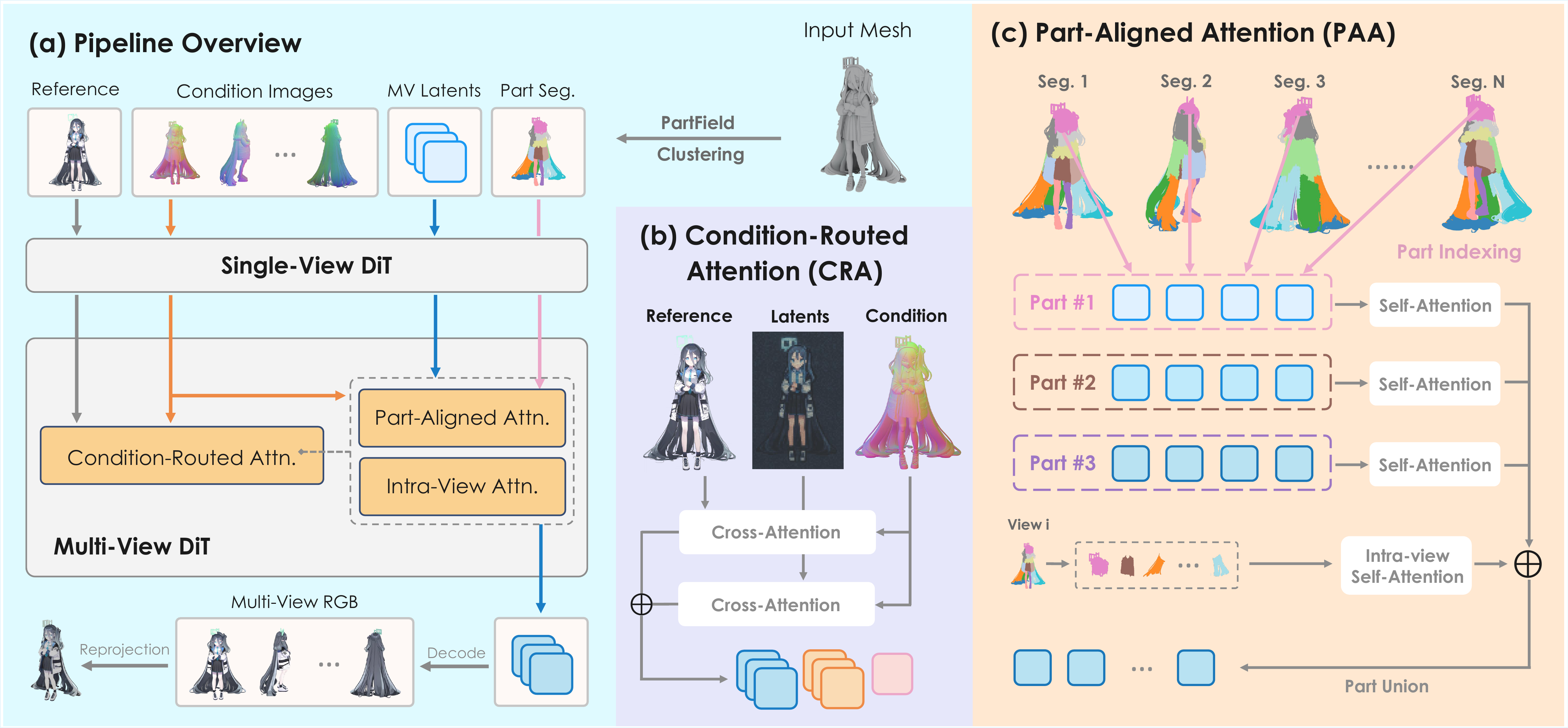}
    \caption{\textbf{Overview of our method.} (a) We employ a two-stage generation framework: the \textit{Single-View DiT} captures intra-view correlations, while the \textit{Multi-View DiT} enhances geometric alignment and cross-view consistency using (b) \textit{Condition-Routed Attention} and (c) \textit{Part-Aligned Attention}. The generated multi-view images are then projected back and inpainted to produce the final 3D texture.}

    \vspace{-0.6em}
    \label{fig:pipeline}
\end{figure*}

% \weikai{We should at least mention how to obtain final texture from multi-view images in method, e.g. how to fuse multi-view image, as we are writing a 3D texture gen. paper, not a NVS paper.}
% done.

\subsection{Overview}
The overview of our method is shown in Fig.~\ref{fig:pipeline}. Following previous works~\cite{liang2025UnitTEX, hunyuan3d2025hunyuan3d2.1}, we finetune a large-scale image generation network (FLUX.1-Kontext~\cite{labs2025flux1kontextflowmatching}) to synthesize six consistent views from an untextured mesh and a reference image.
%which are then projected back into the texture map using a weighted blending mechanism~\cite{bensadoun2024meta, zhao2025hunyuan3d, lu2024genesistex2stableconsistenthighquality}.
% These generated views are then projected back to form the texture map, where an inpainting module is applied to fill occluded regions.
These generated views are then reprojected to form the texture map and inpainting is applied to fill occluded regions.
The backbone of our framework is described in Section~\ref{method3.2}, which consists of two main components: Single-View DiT and Multi-View DiT. In the Single-View DiT stage, we perform batch-wise full attention to capture global information within individual views. In the Multi-View DiT stage, we unfold the batch dimension to perform multi-view attention, enabling information exchange across different views. Our implementations of the modules in Multi-View DiT stage are explained in detail in Section~\ref{method3.3} and Section~\ref{method3.4}. 
\subsection{Multi-View Generation Network}
\label{method3.2}

In this section, we describe the general structure of our multi-view generation network based on Flux~\cite{blackforest2024flux}. 
Given an untextured mesh and a reference image as inputs, we first render normal maps and canonical coordinate maps (CCMs) from the mesh under six predefined viewpoints. These two maps are averaged and then encoded into the latent space via a variational autoencoder (VAE), producing the geometric condition latent $z_{\mathbf{cond}}$. Similarly, the reference image is encoded into $z_{\mathbf{ref}}$, which is replicated six times to match the batch dimension. Let $z_t$ denote the noise latent at timestep $t$. We concatenate the noise, condition, and reference latents along the sequence dimension, forming $\hat{z}_t \in \mathbb{R}^{6 \times 3L \times C}$, where $L$ and $C$ denote the token length per view and the feature dimension respectively, and the six views are tiled along the batch dimension.
Next, a Single-View DiT performs batch-wise full attention to capture intra-view correspondences among the noise, condition, and reference latents.
%:
% \begin{equation}
% \tilde{z}_t = \operatorname{SingleViewDiT}(\hat{z}_t),
% \end{equation}

% The resulting latent sequence is split into image, condition, and reference parts.

% \[
% \tilde{z}_{\mathbf{img}}, \tilde{z}_{\mathbf{cond}}, \tilde{z}_{\mathbf{ref}} = \operatorname{split}(\tilde{z}_t, L).
% \]
% We then average the reference latent across views:
% \begin{equation}
% \bar{z}_{\mathbf{ref}} = \tfrac{1}{6} \sum_{i=1}^{6} \tilde{z}_{\mathbf{ref}}^{(i)}.
% \end{equation}

Then, we flatten the batch dimension and concatenate all per-view noise latents, condition latents and the view-averaged reference latent to form the input of the next stage, denoted as $\tilde{z}_{\mathbf{mv}}\in \mathbb{R}^{1 \times 13L \times C}$. It is then fed into the Multi-View DiT to enhance cross-view consistency.
%:
% \begin{equation}\label{eq:dit_done}
% z'_t = \operatorname{MultiViewDiT}(\tilde{z}_{\mathbf{mv}}).
% \end{equation}

Finally, the noise latent part $z'_{\mathbf{img}}$ is used as the prediction target, and the model is trained via the flow-matching objective
$
\mathcal{L}(\theta) = 
\mathbb{E}_{t, z_0, \epsilon}\!
\left[\|z'_{\mathbf{img}} - (\epsilon - z_0)\|^2\right]
$~\cite{lipman2023flowmatchinggenerativemodeling}.

\subsection{Condition-Routed Attention}
\label{method3.3}

Next, we describe the two modules in the Multi-View DiT stage.
The input to this stage $\tilde{z}_{\mathbf{mv}} \in \mathbb{R}^{1 \times 13L \times C}$ consists of three parts: the noise tokens and condition tokens from six views, along with the reference image tokens.
Previous approaches~\cite{liang2025UnitTEX, bao2025lumitex} simply apply full attention to learn correlations among these modalities.
However, this often leads to a cross-modality ambiguity problem, where noise tokens alternately attend to either the reference image or the geometry condition, lacking stable geometric alignment and producing artifacts in the generated textures.

To address this issue, we redesign the attention mechanism in the Multi-View DiT as a Condition-Routed Attention (CRA), composed of two parallel attention pathways.
The key idea is to calibrate the visual prior from the reference image through the geometry condition, and then let the resulting geometry-aware features guide the noise latents, thereby improving alignment with geometric priors.

Specifically, we form two token groups:  
(1) a \textit{condition–reference} group that fuses geometric and visual priors, and  
(2) a \textit{noise–condition} group that guides the generation toward geometry-aligned outputs.  
We first compute the self-attention within the \textit{condition–reference} group. Let the query, key, and value matrices derived from all tokens in this group be $(Q_\text{c-r}, K_\text{c-r}, V_\text{c-r})$. The attention is computed as:

\begin{equation}
\text{Attn}_{{\text{c-r}}} = \text{Softmax}\!\left(\frac{Q_{\text{c-r}} K_{\text{c-r}}^\top}{\sqrt{d}}\right)V_{\text{c-r}},
\end{equation}
where $d$ denotes the feature dimension. Meanwhile, the attention within the noise–condition group is calculated as $\text{Attn}_{\text{n-c}}$, which will be detailed in Section~\ref{method3.4}. Finally, we merge the outputs from both groups as:
\begin{equation}
\label{eq_1}
\text{Attn}_{\text{CRA}} = \text{Attn}_{\text{n-c}} \cup \text{Attn}_{\text{c-r}}.
\end{equation}
where $\cup$ denotes that the attention between each token pair is computed only once. The specific implementation of how two attention modules are merged is provided in the supplementary material. This dual-pathway attention operates across all 38 attention blocks of the Multi-View DiT. In each block, the visual priors from reference tokens are fused into the geometry condition tokens, which are then used in the subsequent block to guide generation. This mechanism effectively prevents direct reference copying and promotes texture alignment with the underlying 3D surface.

% \subsection{3D Local Attention Based on Semantic Clustering}
\subsection{Part-Aligned Attention}
\label{method3.4}

In this section, we describe the multi-view attention mechanism within the noise–condition branch in Section~\ref{method3.3}. Previous methods simply employ full attention to capture multi-view correspondences, allowing the model to learn from data that similar shapes should correspond to the same regions and thereby yield coherent textures. However, such global modeling often struggles in cases of symmetry or geometric similarity, resulting in inconsistencies across views.

To address this limitation, we introduce a Part-Aligned Attention (PAA) mechanism that leverages geometric priors to group spatially adjacent tokens across different views. Specifically, we adopt PartField~\cite{liu2025partfieldlearning3dfeature} to decompose the 3D mesh $\mathcal{M}$ into $K$ semantic parts (we set $K=20$, which is suitable for most common 3D objects):
\begin{equation}
\mathcal{M} = \{ \mathcal{P}_1, \mathcal{P}_2, \dots, \mathcal{P}_K \},
\end{equation}
where each face $f \in \mathcal{P}_k$ is assigned a part index $k$. We then render the mesh, all faces belonging to the same part $\mathcal{P}_k$ are assigned an identical color across six views, producing part-colored images.

In the latent preparation stage, the model encodes both the condition and reference images into the latent space using a VAE with a down-sampling factor $F$. The resulting feature maps are then partitioned into patches of size $P$, where each token represents an $FP \times FP$ image region. Let $\mathbf{T}^{(v)} = \{ t^{(v)}_1, t^{(v)}_2, \dots, t^{(v)}_N \}$ denote the set of latent tokens extracted from view $v$, 
where each token $t^{(v)}_i \in \mathbb{R}^d$ corresponds to a local image patch on the rendered condition image or its generative result. 
We denote the union of all multi-view tokens as $\mathbf{T} = \bigcup_{v=1}^{6} \mathbf{T}^{(v)} = \{ t_1, t_2, \dots, t_{6N} \}.$

For each token $t_i \in \mathbf{T}$, we determine its semantic part labels based on the previous rendered part-colored images.
Each pixel is assigned a part index $c(p) \in \{1, \dots, K\}$, where $K$ is the total number of semantic parts. If any pixel within the image patch corresponding to $t_i$ belongs to part $k$, 
we assign $t_i$ to group $\mathcal{G}_k$. 
\begin{equation}
\mathcal{G}_k = \{ t_i \mid \exists p \in \text{Patch}(t_i), \; c(p) = k \}, \quad k = 1, \dots, K.
\end{equation}
Notice that it allows a token to be associated with multiple semantic groups if it overlaps with the boundaries of different parts. Then, we perform self-attention jointly on the noise and condition latent tokens within each part group $\mathcal{G}_k$. 
For tokens belonging to the same group, their query, key, and value matrices are denoted as $Q_k, K_k, V_k$, respectively. 
The part-level self-attention is formulated as:
\begin{equation}
\text{Attn}_k(Q,K,V) = \text{Softmax}\!\left(\frac{Q_k K_k^\top}{\sqrt{d}}\right)V_k.
\end{equation}
%where $d$ denotes the feature dimension. 
The outputs from all part groups are then aggregated to obtain the overall part-aligned attention:
\begin{equation}
\label{eq_2}
\text{Attn}_{\text{PAA}}(Q,K,V) = \bigcup_{k=1}^{K} \text{Attn}_k(Q,K,V).
\end{equation}
Also, to maintain global perception within each view, we retain full attention for intra-view context:
\begin{equation}
\text{Attn}_{\text{intra}}^{(v)}(Q,K,V) = \text{Softmax}\!\left(\frac{Q^{(v)} {K^{(v)}}^\top}{\sqrt{d}}\right)V^{(v)},
\end{equation}
where $Q^{(v)}$, $K^{(v)}$, and $V^{(v)}$ denote the query, key, and value matrices derived from tokens of view $v$. Finally, we union the part-aligned and intra-view attention together, obtaining the noise-condtion branch attention:
\begin{equation}
\label{eq_3}
\text{Attn}_{\text{n-c}} = \text{Attn}_{\text{PAA}} \cup \text{Attn}_{\text{intra}}.
\end{equation}
This design constrains cross-view attention to small, semantically coherent 3D-local regions. By aligning attention with 3D semantic parts, PAA strengthens the model’s perception of 3D spatial structure and corrects misaligned correspondences, resulting in more coherent textures.

\section{Experiments}

\definecolor{tabfirst}{RGB}{255,204,204} % Pastel red
\definecolor{tabsecond}{RGB}{255,229,204} % Pastel orange
\definecolor{tabthird}{RGB}{255,255,204} % Pastel yellow

\begin{table*}[t]
\centering
% \vspace{-.2em}
\setlength{\tabcolsep}{3pt}
\resizebox{0.9\textwidth}{!}{
\begin{tabular}{c
                ccccc
                ccc}
\toprule
% & \multicolumn{5}{c}{Textured Mesh Evaluation} & \multicolumn{3}{c}{User Study} \\
& \multicolumn{5}{c}{Quantitative Metrics Evaluation} & \multicolumn{3}{c}{User Study} \\
\cmidrule(lr){2-6} \cmidrule(lr){7-9}
Method
& {FID$\downarrow$} & {CLIP-FID$\downarrow$}  & {CMMD$\downarrow$} & {CLIP-I$\uparrow$} & {LPIPS$\downarrow$}  
& {Qual$\uparrow$} & {GeoAlign$\uparrow$}  & {MV-Cons$\uparrow$} \\ \midrule

Step1X-3D~\citep{li2025step1x}
& 254.1  & 28.23   & 3.914    & 0.8433    & 0.3154 & 2.06 & 2.09 & 2.07 \\

UniTEX~\citep{liang2025UnitTEX}
& 176.2  & 17.19   & 1.156    & 0.8818    & 0.3335 & 3.10 & 3.21 & 3.15\\

MV-Adapter~\citep{huang2025mv}  
& 169.4  & 13.61   & 0.747    & 0.8975    & 0.2939 & 3.02 & 3.03 & 2.88     \\

Hunyuan3D-2.1~\citep{hunyuan3d2025hunyuan3d2.1}
& 167.4  & 16.21   & 1.067    & 0.8867    & 0.3215 & 2.48 & 2.60 & 2.51\\
Ours        
& \textbf{157.8}  & \textbf{12.85}   & \textbf{0.672}    & \textbf{0.9106}    & \textbf{0.2508} & \textbf{4.53} & \textbf{4.47} & \textbf{4.52}  \\ 

\bottomrule
\end{tabular}
}
\caption{\textbf{Quantitative comparison with baseline methods.}  
We evaluate generation fidelity using FID, CLIP-FID, CMMD, CLIP-I, and LPIPS on novel-view renderings, and assess human perceptual quality through a user study rating texture quality (\textit{Qual}), geometric alignment (\textit{GeoAlign}), and multi-view consistency (\textit{MV-Cons}). Our method achieves the best performance across all metrics and human evaluations, demonstrating superior visual fidelity, geometric coherence, and consistency across views.}

\label{tab:benchmark}
\vspace{-.5em}
\end{table*}

% Step1X-3D  & 2.06 & 2.09 & 2.07 \\
% Hunyuan    & 2.48 & 2.60 & 2.51 \\
% MV-Adapter & 3.02 & 3.03 & 2.88 \\
% Unitex     & 3.10 & 3.21 & 3.15 \\
% Ours       & \textbf{4.53} & \textbf{4.47} & \textbf{4.52} \\

% \begin{table}[t]
% \centering
% % \vspace{-.2em}
% %\setlength{\tabcolsep}{3pt}
% \resizebox{\columnwidth}{!}{
% \begin{tabular}{c
%                 ccccc
%                 }
% \toprule
% Method
% & {FID$\downarrow$} & {CLIP-FID$\downarrow$}  & {CMMD$\downarrow$} & {CLIP-I$\uparrow$} & {LPIPS$\downarrow$}  
%  \\ \midrule

% Step1X-3D~\citep{li2025step1x}
% & 254.0783  & 28.2332   & 3.9135    & 0.8433    & 0.3154 \\

% UniTEX~\citep{liang2025UnitTEX}
% & 176.2423  & 17.1888   & 1.1562    & 0.8818    & 0.3335 \\

% MV-Adapter~\citep{huang2024mvadapter}  
% & 169.4190  & 13.6138   & 0.7466    & 0.8975    & 0.2939      \\

% Hunyuan3D-2.1~\citep{hunyuan3d2025hunyuan3d2.1}
% & 167.4491  & 16.2145   & 1.0670    & 0.8867    & 0.3215 \\
% Ours        
% & \textbf{157.7893}  & \textbf{12.8515}   & \textbf{0.6724}    & \textbf{0.9106}    & \textbf{0.2508}   \\ 

% \bottomrule
% \end{tabular}
% }
% \caption{\textbf{Quantitative comparison with baseline methods.}
% We evaluate fidelity on textured mesh renderings using FID, CLIP-FID, CMMD, CLIP-I, and LPIPS.}
% \label{tab:benchmark}
% \vspace{-.5em}
% \end{table}

\subsection{Implementation Details}
% \paragraph{Training.}
We selected 80k objects from Objaverse-XL~\cite{deitke2023objaversexl} and Texverse~\cite{zhang2025texverseuniverse3dobjects} for model training. For each object, we rendered six pre-defined viewpoints at a resolution of 768×768, which were used as the ground truth multi-view images. Additionally, a random view was rendered as the reference image. The back-projection and inpainting modules are adopted from Lumitex~\cite{bao2025lumitex}.
Our DiT backbone was initialized from FLUX.1-Kontext~\cite{labs2025flux1kontextflowmatching}, and we integrate a LoRA adapter~\cite{hu2022lora} with a rank of 16. The model was trained on 8 GPUs for approximately 600 GPU hours in total.
% \paragraph{Inference.}
% During inference, our multi-view generation network synthesizes six views from the given mesh and reference image. These views are then projected back onto the mesh surface using a weighted blending mechanism~\cite{bensadoun2024meta, zhao2025hunyuan3d, lu2024genesistex2stableconsistenthighquality}, yielding the final texture map.
% For texture inpainting, we adopt the LVSM-based inpainting module~\cite{jin2025lvsm} introduced in our previous work LumiTex, to fill in occluded or missing regions.

\subsection{Comparisons}
\paragraph{Baselines.} We compare our method with several open-source state-of-the-art image-to-texture approaches, including MV-Adapter~\cite{huang2025mv}, UniTEX~\cite{liang2025UnitTEX}, Step1X-3D~\cite{li2025step1x}, and Hunyuan3D 2.1~\cite{hunyuan3d2025hunyuan3d2.1}. In addition, we also include comparisons with high-fidelity commercial models, as discussed in Section \ref{qualiative_comp}.
\paragraph{Metrics.}
We use FID~\cite{heusel2017fid}, CLIP-FID, CLIP Maximum-Mean Discrepancy (CMMD)~\cite{jayasumana2024cmmd}, and CLIP-I to evaluate the fidelity of generated textures. We further use LPIPS~\cite{zhang2018lpips} to measure the perceptual similarity between the generated textures and the ground truth.

\begin{figure*}[t]
    \centering
    \includegraphics[width=0.97\linewidth]{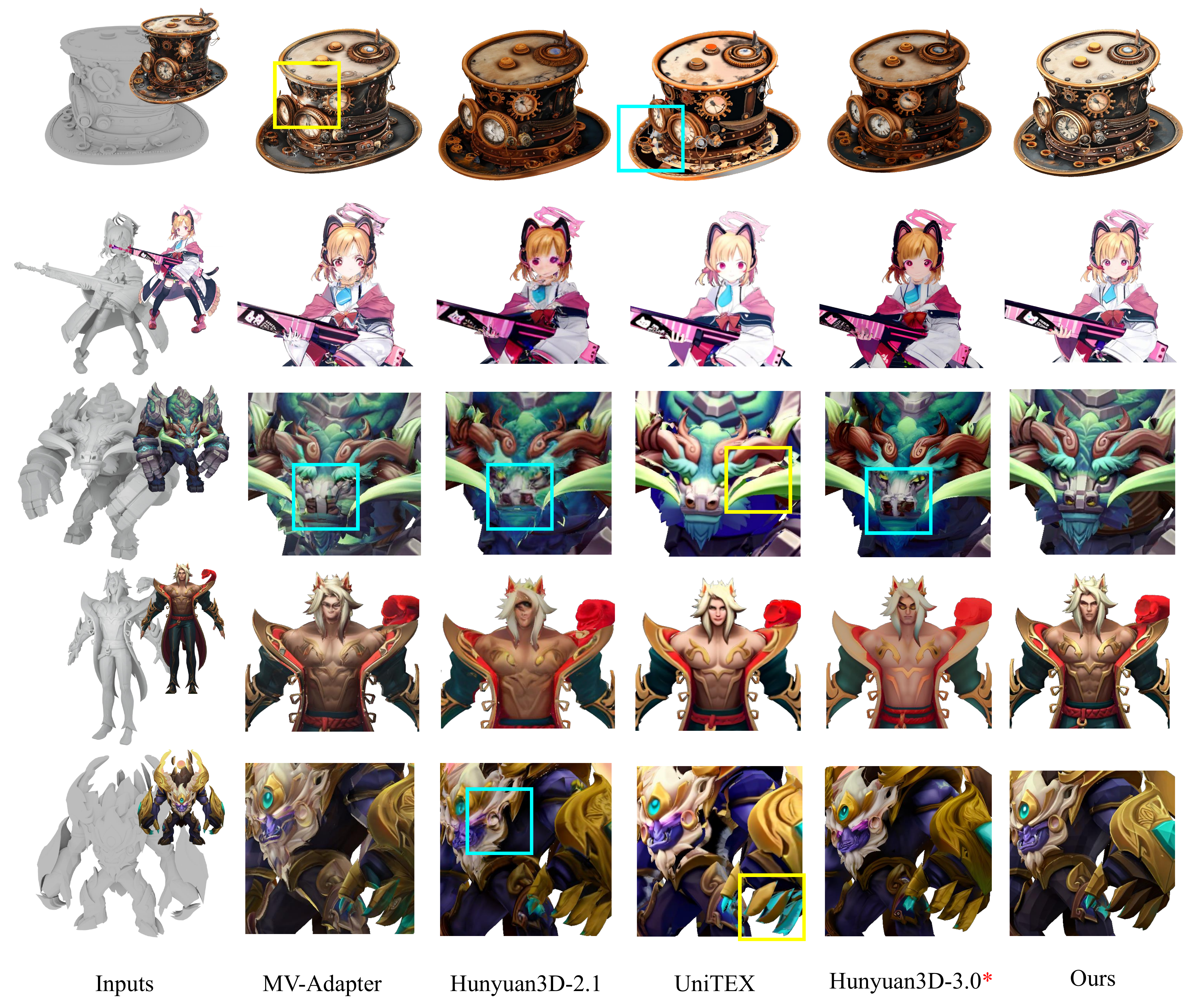}
    \vspace{-0.8em}
    \caption{
    \textbf{Qualitative comparison with recent methods.}
We compare our approach with both open-source and commercial models (marked with \textcolor{red}{*}) on various objects. Regions highlighted in \textcolor[RGB]{245,196,0}{yellow} indicate seams or cross-view inconsistencies, while regions highlighted in \textcolor{cyan}{blue} denote misalignment with the underlying geometry.
Please zoom in for more details. Meshes are generated by Hunyuan3D-3.0.
    }
    \vspace{-1.5em}
    \label{fig:comp}
    \vspace{1em}
\end{figure*}

% \begin{table}[t]
% \centering
% \resizebox{\linewidth}{!}{
% \begin{tabular}{lccc}
% \toprule
% Method & Material Quality & Geometric Consistency & Multi-View Consistency \\
% \midrule
% Step1X-3D  & 2.06 & 2.09 & 2.07 \\
% Hunyuan    & 2.48 & 2.60 & 2.51 \\
% MV-Adapter & 3.02 & 3.03 & 2.88 \\
% Unitex     & 3.10 & 3.21 & 3.15 \\
% Ours       & \textbf{4.53} & \textbf{4.47} & \textbf{4.52} \\
% \bottomrule
% \end{tabular}}
% \caption{\textbf{User study results}. Scores range from 1 to 5.}
% \vspace{-.5em}
% \label{tab:userstudy}
% \end{table}

\subsubsection{Quantiative Comparison}
\label{quantiative_comp}
\paragraph{Textured Mesh Evaluation.}
We evaluate our method on a diverse test set including objects from Objaverse~\cite{deitke2023objaverse} and high-quality game assets, ranging from simple objects to highly detailed models with complex geometry and refined textures. To compare fairly with PBR-based methods, we use illumination-free albedo renderings as reference images. Specifically, for game characters, a random animated pose is sampled as reference while textures are generated on the canonical A-pose to simulate shape discrepancies between reference images and 3D geometries. Each mesh is rendered from 32 viewpoints and compared with the corresponding ground truth albedo images. 

As shown in Tab.~\ref{tab:benchmark}, our method achieves the lowest FID, indicating higher overall fidelity. It also attains superior semantic fidelity, achieving lower CLIP-FID and CMMD and higher CLIP-I, as well as better perceptual quality reflected by lower LPIPS. 

\paragraph{User Study.}
To further assess human perceptual quality, we conduct a user study with 25 participants who rate results (1–5) from different methods along three perceptual dimensions: texture overall quality (\textit{Qual}), geometric alignment (\textit{GeoAlign}), and multi-view consistency (\textit{MV-Cons}). Texture quality measures the visual appearance and fidelity to reference images, geometric alignment evaluates alignment with the underlying mesh, and multi-view consistency checks for naturalness across views, avoiding ghosting and seams. As shown in Tab.~\ref{tab:benchmark}, our method receives the highest scores on the overall quality, showing strong agreement with human preference. Facilitated by our Condition-Routed Attention and Part-Aligned Attention modules, our method also achieves strong performance on geometric alignment and multi-view consistency.

\subsubsection{Qualitative Comparison}
\label{qualiative_comp}
We further present qualitative comparisons with other methods in Fig.~\ref{fig:comp}, including Hunyuan3D-3.0 to compare with the latest commercial models. MV-Adapter, Hunyuan3D-2.1 and Hunyuan3D-3.0 tend to produce blurry textures on geometrically complex regions and sometimes fail to align semantically with the reference image in certain parts. UniTEX generates overly smooth textures, resulting in the loss of fine local details. In addition, its multi-view consistency is relatively low, producing visible seams in the textures. In contrast, our method preserves almost all fine details from the reference images and achieves higher visual fidelity than all other approaches. Furthermore, with our Condition-Routed Attention, we maintain strong geometric coherence, producing minimal artifacts, while our Part-Aligned Attention improves multi-view consistency, resulting in textures that are consistent across views and free of seams.

\subsection{Ablation Study}
\subsubsection{Pixel-Level Multi-View Alignment}

To quantitatively evaluate the effectiveness of our two proposed attention modules, we introduce a metric called \textbf{Multi-View Mean Squared Error (MV-MSE)}, which measures the pixel-level consistency across different rendered views. 
For each pair of generated views among the six pre-defined viewpoints, we compute the average MSE over pixels that correspond to the same 3D locations. 
Formally, MV-MSE is defined as:

\begin{equation}
\begin{split}
\text{MV-MSE} = \frac{2}{N(N-1)} \sum_{(i,j)} \frac{1}{|\Omega_{i}(j)|} \sum_{p \in \Omega_{i}(j)}\\
\| I_i(p) - I_j(\pi_{j}(X_p)) \|_2^2,
\end{split}
\end{equation}
where $I_i$ and $I_j$ denote the generated images from views $i$ and $j$, $X_p$ is the 3D position corresponding to pixel $p$ in view $i$, $\pi_j(\cdot)$ projects the 3D point into view $j$, and $\Omega_i(j)$ denotes the set of pixels in view $i$ whose projected 3D points have a valid corresponding pixel in view $j$ within a small spatial threshold.

% In practice, we first obtain the 3D position of each pixel under both viewpoints using the rendered depth maps and camera parameters. 
% Then, for each pixel in view $i$, we locate its corresponding pixel in view $j$ via projection. 
% If the 3D positional deviation between the two pixels is below a threshold, they are considered to represent the same surface point. 
% The MV-MSE is computed by averaging the MSE over all such pixel pairs across all view pairs.

As shown in Tab.~\ref{tab:mv_align}, our modules reduce MV-MSE by mitigating geometry–texture mismatch and enhancing cross-view consistency.

\begin{table}[t]
\centering
\vspace{4pt}
%\resizebox{\linewidth}{!}{
\begin{tabular}{l c}
\toprule
Method & MV-MSE $\downarrow$ \\
\midrule
Ours & \textbf{0.0384} \\
w/o Part-Aligned & 0.0415 \\
w/o Condition-Routed Attention & 0.0403 \\
\bottomrule
\end{tabular}
%}
\caption{\textbf{Ablation on pixel-level multi-view alignment.}}
\vspace{-1em}
\label{tab:mv_align}
\end{table}

\subsubsection{Part-Aligned Attention}
\begin{figure}
    \centering
    \includegraphics[width=0.95\linewidth]{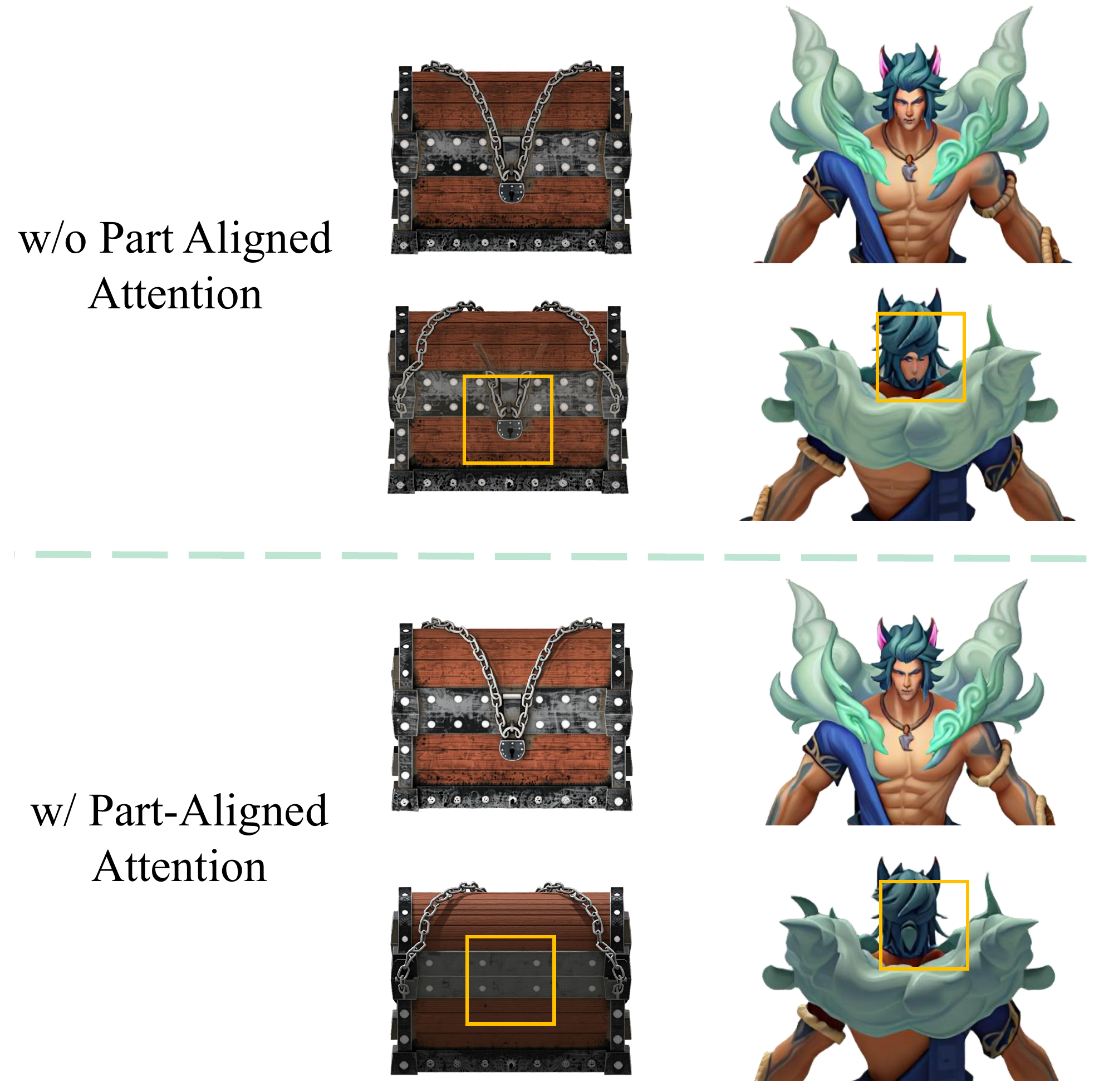}
    \caption{\textbf{Ablation study of Part-Aligned Attention.} Without Part-Aligned Attention, ambiguous cross-view attention causes incorrect alignment across views, while our method yields correct results.}
    \label{fig:ablation_1}
    \vspace{-1em}
\end{figure}

As shown in Fig.~\ref{fig:ablation_1}, removing the Part-Aligned Attention causes the model to ambiguously associate geometrically similar regions as the same location across views. 
Such ambiguous cross-view attention leads to incorrectly aligned multi-view images, producing artifacts in generated 3D textures.
By constraining attention within each local part, the proposed Part-Aligned Attention effectively eliminates these erroneous correspondences, resulting in coherent textures with significantly fewer artifacts across all viewpoints.

\begin{figure}
    \centering
    \includegraphics[width=\linewidth]{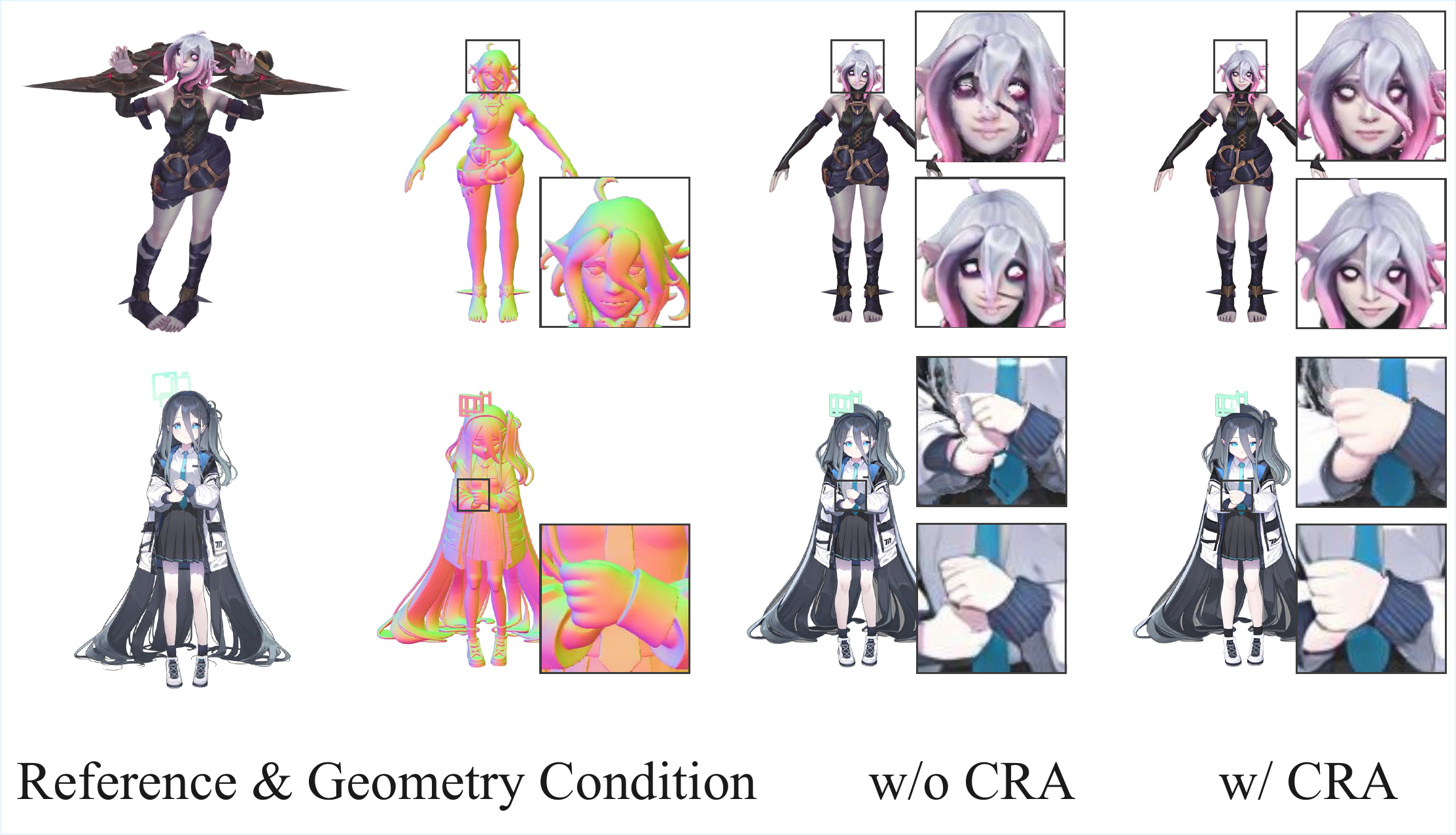}
    \caption{\textbf{Ablation study on the Condition-Routed Attention.}
    We compare textures generated with and without the proposed Condition-Routed Attention.
    The bottom-right inset shows the corresponding multi-view generation results, while the top-right inset illustrates the final textured mesh.}
    \label{fig:ablation_2}
    \vspace{-1em}
\end{figure}

\subsubsection{Condition-Routed Attention}
Fig.~\ref{fig:ablation_2} shows textures generated with and without the Condition-Routed Attention. Removing this module results in inconsistencies between the generated views and the geometry condition. In contrast, by regulating the attention flow through the geometry condition, the module improves texture–geometry alignment and produces more consistent results, mitigating artifacts in the final textures.
\section{Conclusion}
\label{sec:conclusion}
In this work we present \methodName{}, a novel framework for high-fidelity 3D texture generation that explicitly addresses attention ambiguity in multi-view diffusion models. We identify two primary sources of ambiguity: (1) cross-view attention ambiguity among geometrically similar regions, which can cause inconsistent textures across views, and (2) cross-modal attention ambiguity, which can degrade geometry–texture correspondence. To address these problems, we propose two attention mechanisms: Part-Aligned Attention to constrain attention within semantic parts and Condition-Routed Attention to mediate attention flow through the geometry condition. Extensive experiments demonstrate that our method significantly improves texture quality, outperforming existing open-source and commercial baselines. Our study highlights the importance of geometry-aware attention design and takes a step toward automatic, artist-level 3D texture creation.
{
    \small
    \bibliographystyle{ieeenat_fullname}
    \bibliography{main}
}

% WARNING: do not forget to delete the supplementary pages from your submission 
%\input{sec/X_suppl}

\end{document}